\documentclass[journal]{IEEEtran}
\usepackage[utf8]{inputenc}
\usepackage{graphicx}
\usepackage{xcolor}

\begin{document}
\title{A review of deep learning in medical imaging: Imaging traits, technology trends, case studies with progress highlights, and future promises}


\author{S. Kevin Zhou, Hayit Greenspan, Christos Davatzikos, James S. Duncan, Bram van Ginneken, \\Anant Madabhushi, Jerry L. Prince, Daniel Rueckert, Ronald M. Summers
\thanks{S. Kevin Zhou is with School of Biomedical Engineering, University of Science and Technology of China and Institute of Computing Technology, Chinese Academy of Sciences. Hayit Greenspan is with Biomedical Engineering Department, Tel-Aviv University, Israel. Christos Davatzikos is with Radiology Department and Electrical and Systems Engineering Department, University of Pennsylvania, USA. James S. Duncan is with the Departments of Biomedical Engineering and Radiology \& Biomedical Imaging, Yale University. Bram van Ginneken is with Radboud University Medical Center, the Netherlands. Anant Madabhushi with the Department of Biomedical Engineering, Case Western Reserve University and Louis Stokes Cleveland Veterans Administration Medical Center, USA. Jerry L. Prince with  Electrical and Computer Engineering Department, Johns Hopkins University, USA. Daniel Rueckert is with Klinikum rechts der Isar, TU Munich, Germany and Department of Computing, Imperial College, UK. Ronald M. Summers is with National Institute of Health Clinical Center, USA. Zhou and Greenspan are corresponding authors who made equal contributions.}}

\maketitle

\begin{abstract}

Since its renaissance, deep learning has been widely used in various medical imaging tasks and has achieved remarkable success in many medical imaging applications, thereby propelling us into the so-called artificial intelligence (AI) era. It is known that the success of AI is mostly attributed to the availability of big data with annotations for a single task and the advances in high performance computing. However, medical imaging presents unique challenges that confront deep learning approaches. In this survey paper, we first present traits of medical imaging, highlight both clinical needs and technical challenges in medical imaging, and describe how emerging trends in deep learning are addressing these issues. We cover the topics of network architecture, sparse and noisy labels, federating learning, interpretability, uncertainty quantification, etc. Then, we present several case studies that are commonly found in clinical practice, including digital pathology and chest, brain, cardiovascular, and abdominal imaging. Rather than presenting an exhaustive literature survey, we instead describe some prominent research highlights related to these case study applications. We conclude with a discussion and presentation of promising future directions.
\end{abstract}



\section{Overview}

Medical imaging \cite{beutel2000handbook} exploits physical phenomena such as light, electromagnetic radiation, radioactivity, nuclear magnetic resonance, and sound to generate visual representations or images of external or internal tissues of the human body or a part of the human body in a non-invasive manner or via an invasive procedure. The most commonly used imaging modalities in clinical medicine include X-ray radiography, computed tomography (CT), magnetic resonance imaging (MRI), ultrasound, and digital pathology. Imaging data account for about 90\% of all healthcare data\footnote{``The Digital Universe Driving Data Growth in Healthcare,'' published by EMC with research and analysis from IDC (12/13).} and hence is one of the most important sources of evidence for clinical analysis and medical intervention.

\subsection{Traits of medical imaging}  \label{sec.traits}
\begin{figure}[t]
    \centering
    \includegraphics[width=1\columnwidth]{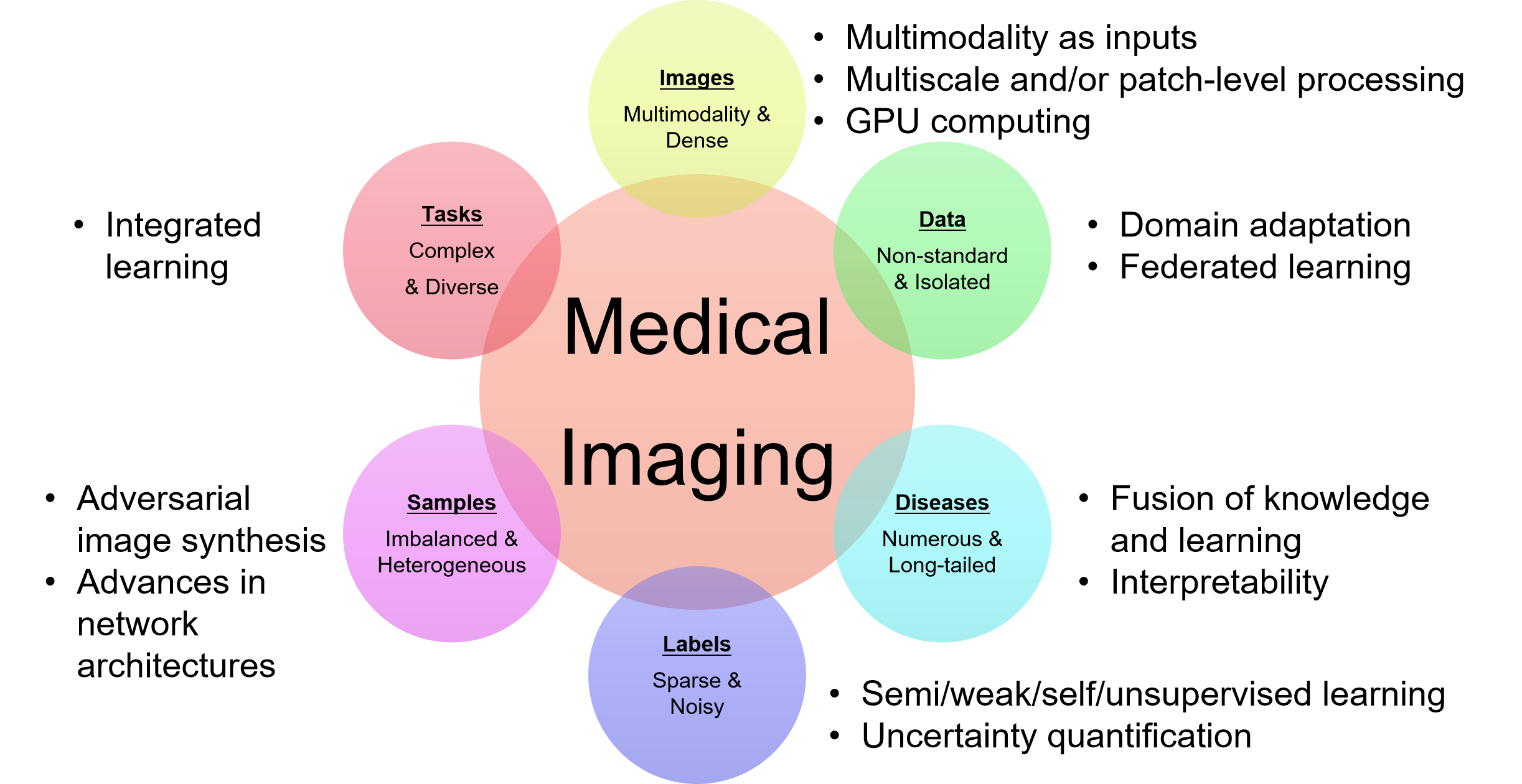}
    \caption{The main traits of medical imaging and the associated technological trends for addressing these traits.}
    \label{fig:traits}
\end{figure}

As described below and illustrated in Figure~\ref{fig:traits}, medical imaging has several traits that influence the suitability and nature of deep learning solutions. {Note that these traits are not necessarily unique to medical imaging. For example, satellite imaging shares the first trait described below with medical imaging.}

\textit{Medical images have multiple modalities and are dense in pixel resolution.} There are many existing imaging modalities and new modalities such as spectral CT are being routinely invented. Even for commonly used imaging modalities, the pixel or voxel resolution has become higher and the information density has increased. For example, the spatial resolution of clinical CT and MRI has reached the sub-millimeter level and the spatial resolution of ultrasound is even better while its temporal resolution exceeds real-time.

\textit{Medical image data are isolated and acquired in non-standard settings.} Although medical imaging data exist in large numbers in the clinic, due to a lack of standardized acquisition protocols there is a large variation in terms of equipment and scanning settings, 
leading to the so-called ``distribution drift'' phenomenon. Due to patient privacy and clinical data management requirements, images are scattered among different hospitals and imaging centers, and truly centralized open source medical big data are rare.

\textit{The disease patterns in medical images are numerous and their incidence exhibits a long tail distribution.} Radiology Gamuts Ontology \cite{budovec2014informatics} defines 12,878 ``symptoms'' (conditions that lead to results) and 4,662 ``diseases'' (imaging findings). The incidence of disease has a typical long-tailed distribution: while a small number of common diseases have sufficient observed cases for large-scale analysis, most diseases are infrequent in the clinic. In addition, novel contagious diseases that are not represented in the current ontology, such as the outbreak of COVID-19, occur with some frequency.

\textit{The labels associated with medical images are sparse and noisy.} Labeling or annotating a medical image is time-consuming and expensive. Also, different tasks require different forms of annotation which creates the phenomenon of label sparsity. Because of variable experience and different conditions, both inter-user and intra-user labeling inconsistency is high \cite{zou2004statistical} and labels must therefore be considered to be noisy. In fact, the establishment of gold standards for image labeling remains an open issue.

\textit{Samples are heterogeneous and imbalanced.} 
In the already labeled images, the appearance varies from sample to sample, with its probability distribution 
being multi-modal. The ratio between positive and negative samples is extremely uneven. For example, the number of pixels belonging to a tumor is usually one to many orders of magnitude less than that of normal tissue.

\textit{Medical image processing and analysis tasks are complex and diverse.} Medical imaging has a rich body of tasks. At the technical level, there is an array of technologies including reconstruction, enhancement, restoration, classification, detection, segmentation, and registration. When these technologies are combined with multiple image modalities and numerous disease types, a very large number of highly-complex tasks associated with numerous applications are formed and should be addressed.

\subsection{Clinical needs and applications} 
Medical imaging is often a key part of the medical diagnosis and treatment process.  Typically, a radiologist reviews the acquired medical images and write a summarizing report of their findings. The referring physician defines a diagnosis and treatment plan based on the images and radiologist's report. Often, medical imaging is ordered as part of a patient's follow-up to verify successful treatment. In addition, images are becoming an important component of invasive procedures, being used both for surgical planning as well as for real-time imaging during the procedure itself. 

As a specific example, we can look at what we term the ``radiology challenge''~\cite{rubin2011informatics,recht2020integrating}. In the past decade, with the development of technologies related to the image acquisition process, imaging devices have improved in speed and resolution. For example, in 1990 a CT scanner might acquire 50--100 slices whereas today's CT scanners might acquire 1000--2500 slices per case. A single whole slide digital pathology image corresponding to a single prostate biopsy core can easily occupy 10GB of space at 40x magnification. Overall, there are billions of medical imaging studies conducted per year, worldwide, and this number is growing.  

Most interpretations of medical images are performed by physicians and, in particular, by radiologists. Image interpretation by humans, however, is limited due to human subjectivity, the large variations across interpreters, and fatigue. Radiologists that review cases have limited time to review an ever-increasing number of images, which leads to missed findings, long turn-around times, and a paucity of numerical results or quantification. This, in turn, drastically limits the medical community's ability to advance towards more evidence-based personalized healthcare. 

AI tools such as deep learning technology can provide support to physicians by automating image analysis, leading to what we can term ``Computational Radiology''~\cite{borgers2012computational,tran2020computational}. Among the automated tools that can be developed are {\em{detection}} of pathological findings, {\em{quantification}} of disease extent, {\em{characterization}} of pathologies (e.g., into benign vs malignant), and assorted software tools that can be broadly characterized as {\em{decision support}}. This technology can also extend physicians' capabilities to include the characterization of three-dimensional and time-varying events, which are often not included in today's radiological reports because of both limited time and limited visualization and quantification tools.

\subsection{Key technologies and deep learning}
Several key technologies arise from the various medical imaging applications, including:~\cite{ beutel2000handbook2,prince2006medical,bankman2008handbook,zhou2019handbook}
\begin{itemize}
    \item \textit{Medical image reconstruction~\cite{wang2019machine},} which aims to form a visual representation (aka an image) from signals acquired by a medical imaging device such as a CT or MRI scanner. Reconstruction of high quality images from low doses and/or fast acquisitions has important clinical implications.
    \item \textit{Medical image enhancement,} which aims to adjust the intensities of an image so that the resultant image is more suitable for display or further analysis. Enhancement methods include denoising, super-resolution, MR bias field correction~\cite{gaillochet2020joint}, and image harmonization ~\cite{dewey2019deepharmony}. Recently, much research has focused on modality translation and synthesis, which can be considered as image enhancement steps.  
    \item \textit{Medical image segmentation~\cite{tajbakhsh2020embracing},} which aims to assign labels to pixels so that the pixels with the same label form a segmented object. Segmentation has numerous applications in clinical quantification, therapy, and surgical planning.  
    \item \textit{Medical image registration~\cite{fu2020deep},} which aims to align the spatial coordinates of one or more images into a common coordinate system. Registration finds wide use in population analysis, longitudinal analysis, and multimodal fusion, and is also commonly used for image segmentation via label transfer.  
    \item \textit{Computer aided detection (CADe) and diagnosis (CADx)~\cite{chan2020computer}.} CADe aims to localize or find a bounding box that contains an object (typically a lesion) of interest. CADx aims to further classify the localized lesion as benign/malignant or one of multiple lesion types.
    \item \textit{Others technologies} include landmark detection~\cite{liu2010search}, image or view recognition~\cite{xu2018less}, automatic report generation~\cite{jing2018automatic}, etc. 
\end{itemize}
In mathematics, the above technologies can be regarded as function approximation methods, which approximate the true mapping $F$ that takes an image (or multiple images if multimodality is accessible) as input and outputs a specific $y$, 
    $y = F(x)$.
The definition of $y$ varies depending on the technology, which itself depends on the application or task. In CADe, $y$ denotes a bounding box. In image registration, $y$ is a deformation field. In image segmentation, $y$ is a label mask. In image enhancement, $y$ is a quality-enhanced image typically of the same size\footnote{Image super-resolution generates an output image that has a different size from the input image.} of the input image $x$. 

There are many ways to approximate $F$; however, deep learning (DL)~\cite{lecun2015deep}, a branch of machine learning (ML), is one of the most powerful methods for function approximation.  Since its renaissance, deep learning has been widely used in various medical imaging tasks and has achieved substantial success in many medical imaging applications. Because of its focus on learning rather than modeling, the use of DL in medical imaging represents a substantial departure from previous approaches in medical imaging. Take supervised deep learning as an example. Assume that a training dataset $\{(x_n,y_n);n=1,\ldots,N\}$ is available and that a deep neural network is parameterized by $\theta$, which includes the number of layers, the number of nodes of each layer, the connecting weights, the choices of activation functions, etc. The neural network that is found to approximate $F$ can be written as  $\phi_{\hat{\theta}}(x)$ where $\hat{\theta}$ are the parameters that minimize the so-called loss function 
\begin{equation}
     L(\theta)= \frac{1}{N} \sum_{n=1}^N l( \phi_\theta(x_n), y_n) + R_1(\phi_\theta(x_n)) + R_2(\theta) \,. 
\end{equation}
Here, $l( \phi_\theta(x), y)$ is the item-wise loss function that penalizes the prediction error, $R_1(\phi_\theta(x_n))$ reflects the prior belief about the output, and $R_2(\theta)$ is a regularization term about the network parameters.  Although the neural network $\phi_{\hat{\theta}}(x)$ does represent a type of model, it is generally thought of as a ``black box'' since it does not represent a designed model based on well-understood physical or mathematical principles.  


There are many survey papers on deep learning based key technologies for medical image analysis~\cite{greenspan2016guest,litjens2017survey,zhou2017deep,shen2017deep,ker2017deep,yi2019generative,cheplygina2019not,hesamian2019deep,duncan2019biomedical,haskins2020deep}.  To differentiate the present review paper from these works, we specifically omit any presentation of the technical details of DL itself, which is no longer considered new and is well-covered in numerous other works, and focus instead on the connections between the emerging DL approaches and the specific needs in medical imaging and on several case examples that illustrate the state of the art.  


\subsection{Historical perspective}
Here, we briefly outline the development timeline of DL in medical imaging.  Deep learning was termed one of the 10 breakthrough technologies of 2013 \cite{MITTech_2013}. This followed the 2012 large-scale image categorization challenge that introduced the  CNN superiority on the ImageNet dataset \cite{krizhevsky2012imagenet}. At that time, DL emerged as the leading machine-learning tool in the general imaging and computer vision domains and the medical imaging community began a debate about whether DL would be applicable in the medical imaging space.  The concerns were due to the challenges we have outlined above, with the main challenge being the lack of sufficient labeled data, known as the {\em{data challenge}}.

Several steps can be pointed to as {\em{enablers}} of the DL technology within the medical imaging space: In 2015--2016, techniques were developed using “transfer learning” (TL) (or what was also called “learning from non-medical features” \cite{Bar_TL_2015})  to apply the knowledge gained via  solving  a  source  problem  to  a  different  but  related  target problem.  A key question was whether a network pre-trained on natural imagery would be applicable to medical images. Several groups showed this to be the case (e.g. \cite{shin2016deep,gulshan2016development,Bar_TL_2015}); using  the deep  network  trained  based  on  ImageNet  and  fine-tuning  to  a  medical  imaging  task  was helpful in  order  to  speed  up  training convergence  and  improve  accuracy.  

In 2017--2018, {\em {synthetic data augmentation}} emerged as a second solution to process limited datasets. Classical augmentation is a key component of any network training. Still, key questions to address were whether it was possible to synthesize {{medical data}} using schemes such as generative modeling and whether the synthesized data would serve as viable medical examples and would in practice increase performance of the medical task at hand. Several works across varying domains demonstrated that this was in fact the case. In ~\cite{Maayan_GAN_Lesion_2018}, for example, synthetic image augmentation based on generative adversarial network (GAN) was shown to generate lesion image samples that were not recognized as synthetic by the expert radiologists and also increased CNN performance in classifying liver lesions. 
GANs, variational encoders, and variations on these are still being explored and advanced in recent works, as will be described in the following Section.

For image segmentation, one of the key contributions that emerged from the medical imaging community was the U-Net architecture \cite{ronneberger2015u}.  Originally designed for microscopic cell segmentation, the U-Net has proven to efficiently and robustly learn effective features for many medical image segmentation tasks.

 
\subsection{Emerging deep learning approaches}

\underline{Network architectures.} Deep neural networks have a larger model capacity and stronger generalization capability than shallow neural networks. Deep models trained on large scale annotated databases for a single task achieve outstanding performances, far beyond traditional algorithms or even human capability.

\textit{Making it deeper.} Starting from AlexNet~\cite{krizhevsky2012imagenet}, there was a research trend to make networks deeper, as represented by VGGNet~\cite{simonyan2014very}, Inception Net~\cite{szegedy2015going}, and ResNet~\cite{he2016deep}.
The use of skip connections makes a deep network more trainable as in DenseNet~\cite{huang2017densely} and U-Net~\cite{ronneberger2015u}. U-net was first proposed to tackle segmentation while the other networks were developed for image classification. Deep supervision~
\cite{lee2015deeply} further improves discriminative power.

\textit{Adversarial and attention mechanisms}. In the generative adversarial network (GAN) ~\cite{goodfellow2014generative}, Goodfellow et al. propose to accompany a generative model with a discriminator that tells whether a sample is from the model distribution or the data distribution. Both the generator and discriminator are represented by deep networks and their training is done via a minimax optimization. Adversarial learning is widely used in medical imaging~\cite{yi2019generative}, including medical image reconstruction~\cite{wang2019machine}, image quality enhancement~\cite{dewey2019deepharmony}, and segmentation~\cite{yang2017automatic}.  

Attention mechanism~\cite{xu2015show} allows automatic discovery of ``where'' and ``what'' to focus on when describing image contents or making a holistic decision. Squeeze and excitation~\cite{hu2018squeeze} can be regarded as a channel attention mechanism. Attention is combined with GAN in~\cite{zhang2019self} and with U-Net in~\cite{oktay2018attention}.

\textit{Neural architecture search (NAS) and light weight design}. NAS ~\cite{elsken2018neural} aims to automatically design the architecture of a deep network for high performance geared toward a given task. Zhu et al.~\cite{zhu2019vnas} successfully apply NAS to volumetric medical image segmentation. Light weight design~\cite{howard2017mobilenets,zhang2018shufflenet}, on the other hand, aims to design the architecture for computational efficiency on resource-constrained devices such as mobile phones while maintaining accuracy.


\underline{Annotation efficient approaches.} To address sparse and noisy labels, we need DL approaches that are efficient with respect to annotations. So, a key idea is to leverage the power and robustness of feature representation capability derived from existing models and data even though the models or data are not necessarily from the same domain or for the same task and to adapt such representation to the task at hand. To do this, there are a handful of methods proposed in the literature~\cite{cheplygina2019not}, including transfer learning, domain adaptation, self-supervised learning, semi-supervised learning, weakly/partially supervised learning, etc.

\textit{Transfer learning (TL)} aims to apply the knowledge gained via solving a source problem to a different but related target problem. One commonly used TL method is to use the deep network trained based on ImageNet and fine tune it to a medical imaging task in order to speed up training convergence and improve accuracy. With the availability of a large number of annotated datasets, such TL methods~\cite{gulshan2016development} achieve remarkable success. However, ImageNet consists of natural images and its pretrained models are for 2D images only and are not necessarily the best for medical images, especially for small-sample settings~\cite{raghu2019transfusion}. Liu et al.  \cite{liu20183d} propose a 3D anisotropic hybrid network that effectively transfers convolutional features learned from 2D images to 3D anisotropic volumes. In~\cite{chen2019med3d}, Chen et al. combine multiple datasets from several medical challenges with diverse modalities, target organs, and pathologies and learn one 3D network that provides an effective pretrained model for 3D medical image analysis tasks. 

\textit{Domain adaptation} is a form of transfer learning in which the source and target domains have the same feature space but different distributions. In~\cite{kamnitsas2017unsupervised}, domain-invariant features are learned via an adversarial mechanism that attempts to classify the domain of the input data. Zhang et al.~\cite{zhang2018translating} propose to synthesize and segment multimodal medical volumes using generative adversarial networks with cycle- and shape-consistency. In~\cite{dou2018unsupervised}, a domain adaptation module that maps the target input to features which are aligned with source domain feature space is proposed for cross-modality
biomedical image segmentation, using a domain critic module for discriminating the feature space of both domains. Huang et al.~\cite{huang20193d} propose a universal U-Net comprising domain-general and domain-specific parameters to deal with multiple organ segmentation tasks on multiple domains. This \textit{integrated learning} mechanism offers a new possibility of dealing with multiple domains and even multiple heterogeneous tasks. 

\textit{Self-supervised learning}, a form of unsupervised learning, learns a representation through a proxy task, in which the data provides supervisory signals. Once the representation is learned, it is fine tuned by using annotated data.
The models genesis method~\cite{zhou2019models} uses a proxy task of recovering the original image using a distorted image as input. The possible distortions include non-linear gray-value transformation, local pixel shuffling, and image out-painting and in-painting. In~\cite{zhu2020rubik}, Zhu et al. propose solving a Rubik's Cube proxy task that involves three operations, namely cube ordering, cube rotating, and cube masking. This allows the network to learn features that are invariant to translation and rotation and robust to noise as well.

\textit{Semi-supervised learning} often trains a model using a small set of annotated images, then generates pseudo-labels for a large set of images with annotations, and learns a final model by mixing up both sets of images. Bai et al.~\cite{bai2017semi} implement such a method for cardiac MR segmentation. In~\cite{nie2018asdnet}, Nie et al. propose an attention based semi-supervised deep network for segmentation. It adversarially trains a segmentation network, from which a confidence map is computed as a region-attention based semi-supervised learning strategy to include the unlabeled data for training.

\textit{Weakly or partially supervised learning.} 
In~\cite{wang2017chestx}, Wang et al. solve a weakly-supervised multi-label disease classification from a chest x-ray. To relax the stringent pixel-wise annotation for image segmentation, weakly supervised methods that use image-level annotations~\cite{xu2014weakly}, or weak annotations like dots and scribbles~\cite{kervadec2019constrained} are proposed. For multi-organ segmentation, Shi et al.~\cite{shi2020marginal} learn a single multiclass network from a union of multiple datasets, each with a low sample size and partial organ label, using newly proposed marginal loss and exclusion loss. Schleg et al.~\cite{schlegl2019f} build a deep model from only normal images to detect abnormal regions in a test image.

\textit{Unsupervised learning and disentanglement}.  
Unsupervised learning does not rely on the existence of annotated images. A disentangled network structure is designed with an adversarial learning strategy that promotes the statistical matching of deep features has been widely used. In medical imaging, unsupervised learning and disentanglement have been used in image registration~\cite{qin2019unsupervised}, motion tracking~\cite{qin2018joint}, artifact reduction~\cite{liao2019adn},  improving classification~\cite{Avi_IEEEEMBS_2019}, domain adaptation~\cite{Yang_Duncan_2019}, and general modeling~\cite{tsafratis_2019}.

\begin{figure}[t]
    \centering
    \includegraphics[width=\columnwidth]{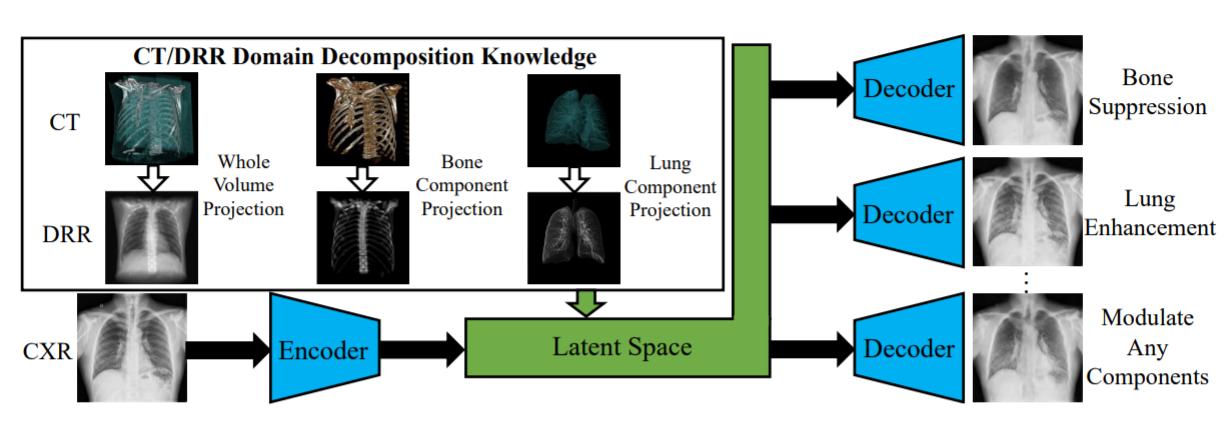}
    \caption{Leveraging the anatomy knowledge embedded in CT to decompose a chest x-ray~\cite{li2019encoding}.}
    \label{fig:xray}
\end{figure}

\underline{Embedding knowledge into learning.}
Knowledge arises from various sources such as imaging physics, statistical constraints, and task specifics and ways of embedding into a DL approach vary too. For chest x-ray disease classification, Li et al.~\cite{li2019encoding,li2020high} encode anatomy knowledge embedded in unpaired CT into a deep network that decomposes a chest x-ray into lung, bone and the remaining structures (see Fig.~\ref{fig:xray}). With augmented bone-suppressed images, classification performance is improved in predicting 11 out of 14 common lung diseases. In \cite{Gozes_2018} lung radiographs are enhanced  by learning to extract lung structures from CT based simulated x-ray (DRRs) and fusing  with the original x-ray image. The enhancement is shown to augment results of pathology characterization in real x-ray images. In \cite{lin2019dudonet}, a dual-domain network is proposed to reduce metal artifacts on both the image and sinogram domains, which are seemingly integrated into one differential framework through a Radon inverse layer, rather than two separate modules.

\underline{Federated learning.} To combat issues related to data privacy, data security, and data access rights, it has become increasingly important to have the capability of learning a common, robust algorithmic model through distributed computing and model aggregation strategies so that no data are transferred outside a hospital or an imaging lab. This research direction is called federated learning (FL) \cite{yang2019federated}, which is in contrast to conventional centralized learning with all the local datasets uploaded to one server. There are many ongoing research challenges related to federate learning such as reduced communication burden~\cite{konevcny2016federated}, data heterogeneity in various local sites~\cite{zhao2018federated}, and vulnerability to attacks~\cite{bagdasaryan2020backdoor}.

Despite its importance, work on FL in medical imaging has only been reported recently. Sheller et al. \cite{sheller2018multi} present the first use of FL for multi-institutional DL model without sharing patient data and report similar brain lesion segmentation performances between the models trained in a federated or centralized way. In \cite{li2019privacy}, Li et al. study several practical FL methods while protecting data privacy for brain tumour segmentation on the BraTS dataset and demonstrate a trade-off between model performance and privacy protection costs. Recently, FL is applied, together with domain adaptation, to train a model with boosted analysis performance and reliable discovery of disease-related biomarkers~\cite{li2020multi}.

\underline{Interpretability.} Clinical decision-making relies heavily on evidence gathering and interpretation. Lacking evidence and interpretation makes it difficult for physicians to trust the ML model's prediction, especially in disease diagnosis. In addition, interpretability is also the source of new knowledge. Murdoch et al. \cite{murdoch2019interpretable} define interpretable machine learning as leveraging machine-learning models to extract relevant knowledge about domain relationships contained in data, aiming to provide insights for a user into a chosen domain problem. Most interpretation methods are categorized as model-based and post-hoc interpretability. The former is about constraining the model so that it readily provides useful information (such as sparsity, modularity, etc.) about the uncovered relationships. The latter is about extracting information about what relationships the model has learned.

\textit{Model-based interpretability.} For cardiac MRI classification~\cite{clough2019global}, diagnostically meaningful concepts in the latent space are encoded. In~\cite{biffi2018learning}, when training the model for healthy and hypertrophic cardiomyopathy classification, it leverages interpretable task-specific anatomic patterns learned from 3D segmentations.

\textit{Post-hoc interpretability.} In ~\cite{li2019graph}, the feature importance scores are calculated for graph neural network by comparing its interpretation ability with Random Forest. Li et al.~\cite{li2018brain} propose a brain biomarker interpretation method through a frequency-normalized sampling strategy to corrupt an image. In~\cite{wickstrom2020uncertainty}, various interpretability methods are evaluated in the context of semantic segmentation of polyps from colonoscopy images.
In~\cite{pereira2018enhancing}, a hybrid RBM-Random Forest system on brain lesion segmentation is learned with the goal of enhancing interpretability of automatically extracted features.

\underline{Uncertainty quantification} characterizes the model prediction with confidence measure~\cite{gal2016dropout}, which can be regarded as a method of post-hoc interpretability, even though often the uncertainty measure is calculated along with the model prediction. Recently there are emerging works that quantify uncertainty in deep learning methods for medical image segmentation~\cite{baumgartner2019phiseg,awate2019estimating,jungo2019assessing}, lesion detection~\cite{nair2020exploring}, chest x-ray disease classification~\cite{ghesu2019quantifying}, and diabetic retinopathy grading~\cite{ayhan2020expert,araujo2020dr}. One additional extension to uncertainty is its combination with the knowledge that the given labels are noisy. Works are now starting to emerge that take into account label uncertainty in the modeling of the network architecture and its training~\cite{dgani2018training}. 


\section{Case Studies with Progress Highlights} 

Given that DL has been used in a vast number of medical imaging applications, it is nearly infeasible to cover all possible related literature in a single paper. Therefore, we cover several selected cases that are commonly found in clinical practices, which include chest, neuro, cardiovascular, abdominal, and microscopy imaging. Further, rather than presenting an exhaustive literature survey for each studied case, we provide some prominent progress highlights in each case study.

\subsection{Deep learning in thoracic imaging} 

Lung diseases have a high mortality and morbidity. In the top ten causes of death worldwide we find lung cancer, chronic obstructive pulmonary disease (COPD), pneumonia, and tuberculosis (TB). At the moment of writing this overview, COVID-19 has a death rate comparable to TB. Imaging is highly relevant to diagnose, plan treatment and learn more about the causes and mechanisms underlying these and other lung diseases. Next to that, pulmonary complications are common in hospitalized patients. As a result, chest radiography is by far the most common radiological examination, often comprising over a third of all studies in a radiology department.

Plain radiography and computed tomography are the two most common modalities to image the chest. The high contrast in density between air-filled lung parenchyma and tissue makes CT ideal for in vivo analysis of the lungs, obtaining high-quality and high-resolution images even at very low radiation dose. Nuclear imaging (PET or PET/CT) is used for diagnosing and staging of oncology patients. MRI is somewhat limited  in the lungs, but can yield unique functional information. Ultrasound imaging is also difficult because sound waves reflect strongly at boundaries of air and tissue, but point-of-care ultrasound is used at the emergency department and is widely used to monitor COVID-19 patients for which the first decision support applications based on deep learning have already appeared~\cite{roy2020deep}.

\underline{Segmentation of anatomical structures.} For analysis and quantification from chest CT scans, automated segmentation of major anatomical structures is an important prerequisite. Recent publications demonstrate convincingly that deep learning is now the state-of-the-art method to achieve this. This is evident from inspecting the results of LOLA11\footnote{https://lola11.grand-challenge.org/}, a competition started in 2011 for lung and lobe segmentation in chest CT. The test dataset for this challenge include many challenging cases with lungs affected by severe abnormalities. For years, the best results were obtained by interactive methods. In 2019 and 2020, seven fully automatic methods based on U-Nets or variants thereof made the top 10 for lung segmentation, and for lobe segmentation, two recent methods obtained results outperforming the best interactive methods. Both these methods~\cite{gerard2019pulmonary,xie2020relational} were trained on thousands of CT scans from the COPDGene study~\cite{regan2010genetic}, illustrating the importance of large high quality datasets to obtain good results with deep learning. This data is publicly available on request. Both methods use a multi-resolution U-Net like architecture with several customizations. Gerard et al. integrate a previously developed method for finding the fissures~\cite{gerard2019fissurenet}. Xie et al.~\cite{xie2020relational} add a non-local module with self-attention and fine-tune their method on data of COVID-19 suspects to accurately segment the lobes in scans affected by ground-glass and consolidations.

Segmentation of the vasculature, separated into arteries and veins, and the airway tree, including labeling of the branches and segmentation of the bronchial wall, is another important area of research. Although methods that use convolutional networks in some of their steps have been proposed, developing an architecture entirely based on deep learning that can accurately track and segment intertwined tree structures and take advantage of the known geometry of these complex structures, is still an open challenge.  

\begin{figure}[t]
    \centering
    \includegraphics[width=\columnwidth]{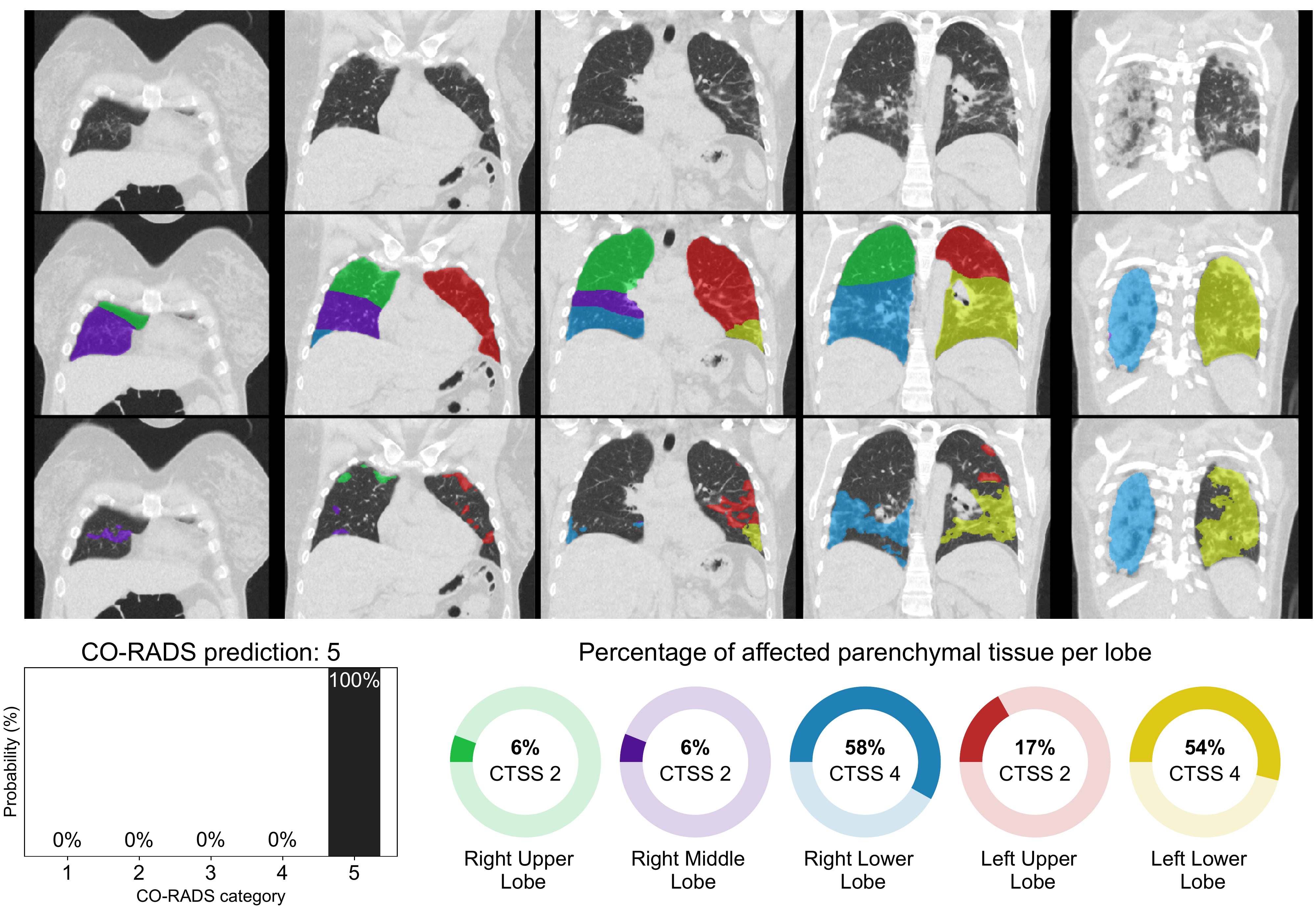}
    \caption{Example output of the CORADS-AI system for a COVID-19 case. Top row shows coronal slices, the second row shows lobe segmentation and bottom row shows detected abnormal areas of patchy ground-glass and consolidation typical for COVID-19 infection. The CO-RADS prediction and CT severity score per lobe are displayed below the images.}
    \label{fig:covid}
\end{figure}

\underline{Detection and diagnosis in chest radiography.} Recently the number of publications on detecting abnormalities in the ubiquitous chest x-ray has increased enormously. 
This trend has been driven by the availability of large public datasets, such as ChestXRay14~\cite{wang2017chestx}, CheXpert~\cite{irvin2019chexpert}, MIMIC~\cite{johnson2019mimic}, and PadChest~\cite{bustos2019padchest}, totaling 868K images. Labels for presence or absence of over 150 different abnormal signs were gathered by text-mining the accompanying radiology reports. This makes the labels noisy. Most publications use a standard approach of inputting the entire image in a popular convolutional network architecture. Methodological contributions include novel ways of preprocessing the images, handling the label uncertainty and the large number of classes, suppressing the bones~\cite{li2020high}, and exploiting self-supervised learning as a way of pretraining. So far, only few publications analyze multiple exams of the same patient to detect interval change or analyze the lateral views. 


\underline{Decision support in lung cancer screening.} Following the positive outcome of the NLST trial, the United States has started a screening program for heavy smokers for early detection of lung cancer with annual low-dose CT scans. Many other countries worldwide are expected to follow suit. In the United States, screening centers have to use a reporting system called Lung-RADS~\cite{Lungrads}. Reading lung cancer screening CT scans is time consuming and therefore automating the various steps in Lung-RADS has received a lot of attention.  

The most widely studied topic is nodule detection~\cite{liu2020evolving}. Nodules may represent lung cancer. Many DL systems were compared in the LUNA16 challenge\footnote{https://luna16.grand-challenge.org/}. Lung-RADS classifies scans in categories based on the most suspicious nodule, and this is determined by the nodule type and size. DL systems to determine nodule type have been proposed~\cite{ciompi2017towards} and measuring the size can be done by traditional methods based on thresholding and mathematical morphology but also with DL networks. Finally, Lung-RADS contains the option to directly refer scans with a nodule that is highly suspicious for cancer. Many DL systems to estimate nodule malignancy have been proposed.

The advantage of automating the LUNG-RADS guidelines step-by-step is that this leads to an explainable AI solution that can directly support radiologists in their reading workflow. Alternatively, one could ask a computer to directly predict if a CT scan contains an actionable lung cancer. This was the topic of a Kaggle challenge organized in 2017\footnote{https://www.kaggle.com/c/data- science-bowl-2017/} in which almost 2000 teams competed for a one million dollar prize. The top 10 solutions all used deep learning and are open source. Two years later, a team from Google published an implementation~\cite{ardila2019end} following the approach of the winning team in the Kaggle challenge, employing modern architectures such as a 3D inflated Inception architecture (I3D)~\cite{carreira2017quo}. The I3D architecture builds upon the Inception v1 model for 2D image classification but inflates the filters and pooling kernels into 3D. This enables the use of an image classification model pre-trained with 2D data for a 3D image classification task. The paper showed that the model outperformed six radiologists that followed Lung-RADS. The model was also extended to handle follow-up scans where it obtained performance slightly below human experts.      


\underline{COVID-19 case study.} As an illustration how DL with pre-trained elements can be used to rapidly build applications, we briefly discuss the development of two tools for COVID-19 detection, for chest radiographs and chest CT. In March 2020, many European hospitals were overwhelmed by patients presenting at the emergency care with respiratory complaints. There was a shortage of molecular testing capacity for COVID-19 and turnaround time for test results was often days. Hospitals therefore used chest x-ray or CT to obtain a working diagnosis and decide whether to hospitalize patients and how to treat them. In just six weeks, researchers from various Dutch and German hospitals, research institutes and a company managed to create a solution for COVID-19 detection from an x-ray and from a CT scan. Figure~\ref{fig:covid} shows the result of this CORADS-AI system for a COVID-19 positive case.

The x-ray solution started from a convolutional network using local and global labels, pretrained to detect tuberculosis~\cite{murphy2020computer}, fine-tuned using public and private data of patients with and without pneumonia to detect pneumonia in general, and subsequently fine-tuned on x-ray data from patients of a Dutch hospital in a COVID-19 hotspot. The system was subsequently evaluated on 454 chest radiographs from another Dutch hospital and shown to perform comparably to six chest radiologists \cite{murphy2020covid}. The system is currently being field tested in Africa.

The CT solution, called CO-RADS~\cite{prokop2020corads}, aimed to automate a clinical reporting system for CT of COVID-19 suspects. This system assesses the likelihood of COVID-19 infection on a scale from CO-RADS 1 (highly unlikely) to CO-RADS 5 (highly likely) and quantifies the severity of disease using a score per lung lobe from 0 to 5 depending on percentage affected lung parenchyma for a maximum CT severity score of 25 points. The previously mentioned lobe segmentation~\cite{xie2020relational} was employed. Abnormal areas in the lung were segmented using a 3D U-net built with the nnU-Net framework~\cite{isensee2019automated} in a cross-validated fashion with 108 scans and corresponding reference delineations to segment ground-glass opacities and consolidation in the lungs. The CT severity score was derived from the segmentation results by computing the percentage of affected parenchymal tissue per lobe. nnU-Net was compared with several other approaches and performed best. For assessing the CO-RADS score, the previously mentioned I3D architecture~\cite{carreira2017quo} performed best. 

\subsection{Deep learning in neuroimaging} 

In recent years, deep learning has seen a dramatic rise in popularity within the neuroimaging community. Many neuroimaging tasks including segmentation, registration, and prediction now have deep learning based implementations. Additionally, through the use of deep generative models and adversarial training, deep learning has enabled new avenues of research in complex image synthesis tasks. With the increasing availability of large and diverse pooled neuroimaging studies, deep learning offers interesting prospects for improving accuracy and generalizability while reducing inference time and the need for complex preprocessing. CNNs, in particular, have allowed for efficient network parameterization and spatial invariance, both of which are critical when dealing with high-dimensional neuroimaging data. The learnable feature reduction and selection capabilities of CNNs have proven effective in high level prediction and analysis tasks and has reduced the need for highly specific domain knowledge. Specialized networks such as U-Nets~\cite{ronneberger2015u}, V-Nets ~\cite{milletari2016v}, and GANs~\cite{goodfellow2014generative} are also popular in neuroimaging, and have been leveraged for a variety of segmentation and synthesis tasks. 

\begin{figure}
    \centering
    \includegraphics[width=0.8\columnwidth]{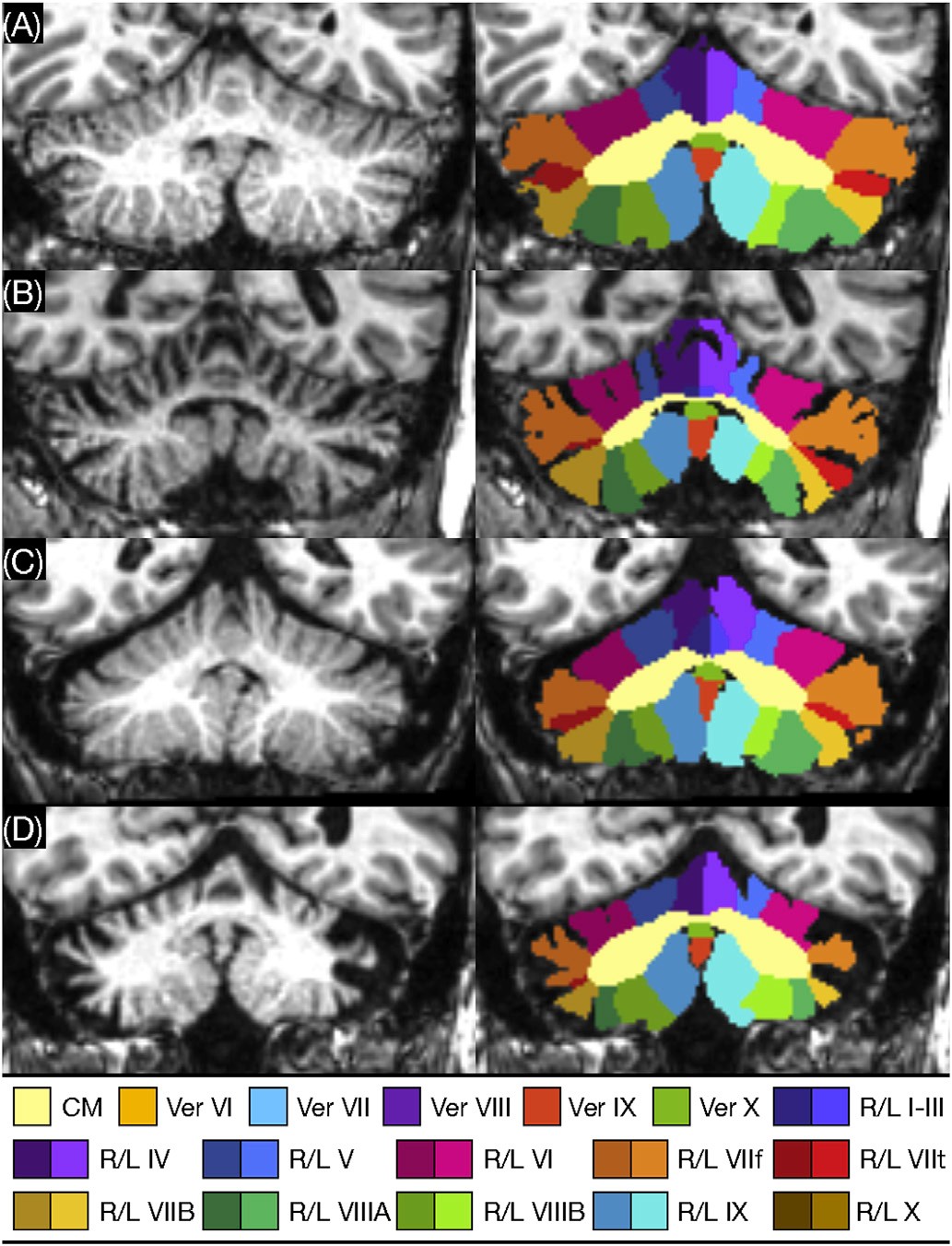}
    \caption{Cerebellum parcellation by the ACAPULCO cascaded deep networks method. Lobule labels are shown for (A) a healthy subject and subjects with (B)~spinocerebellar ataxia (SCA) type 2, (C) SCA type 3, and (D) SCA type 6~\cite{han2020automatic}.}
    \label{fig:ACAPULCO}
\end{figure}

\underline{Neuroimage segmentation and tissue classification}. Accurate segmentation is an important preprocessing step that informs much of the downstream analytic and predictive tasks done in neuroimaging. Commonly used tools such as FreeSurfer~\cite{fischl2012freesurfer} rely on atlas based methods, whereby an atlas is deformably registered to the scan, which requires time consuming optimization problems to be solved. Proposed deep learning based approaches, however, are relatively computationally inexpensive during inference. Recent research has focused on important tasks such as deep learning based brain extraction~\cite{kleesiek2016deep}, cortical and subcortical segmentation~\cite{dolz20183d,chen2018voxresnet,wachinger2018deepnat,huo20193d}, and tumor and lesion segmentation ~\cite{havaei2017brain,kamnitsas2017efficient}. Some interesting research has looked at improving the generalization performance of  deep learning based segmentation methods across neuroimaging datasets imaged at different scanners. In particular, Kamnitsas et al.~\cite{kamnitsas2017unsupervised} have proposed a training schema, which leverages adversarial training to learn scanner invariant feature representations. They use an adversarial network to classify the origin of the input data based on the downstream feature representation learned by the segmentation network. By penalizing the segmentation network for improved performance of the adversarial network, they show improved segmentation generalization across datasets. Brain tumor segmentation has been another active area of research in the neuroimaging community where deep learning has shown promise. In the past, brain tumor datasets have remained relatively small, particularly ones with subjects imaged at  a single institution. The Brain Tumor Segmentation Challenge (BraTS)~\cite{bakas2018identifying} has provided the community with an accessible dataset as well as a way to benchmark various approaches against one another. While it has been seen that deep learning has difficulty in training on datasets with relatively few scans, new architectures and training methods are becoming increasingly effective at this. Havaei et al.~\cite{havaei2017brain} demonstrate the performance of their glioblastoma segmentation network on the BraTS dataset, achieving high accuracy while being substantially faster than previous methods. Another task where deep networks are finding increasing success is semantic segmentation in which anatomical labels are not necessarily well-defined by image intensity changes but can be identified by relative anatomical locations. A good example is that of cerebellum parcellation where deep networks performed best in a recent comparison of methods~\cite{carass2018comparing}. The even newer ACAPULCO method~\cite{han2020automatic} uses two cascaded deep networks to produce cerebellar lobule labels, as shown in Figure~\ref{fig:ACAPULCO}.

\underline{Deformable image registration}.
Image registration allows for imaging analysis in a single subject across imaging modalities and time points. Deep learning based deformable registration with neuroimaging data has proven to be a difficult problem, especially considering the lack of ground truth. Still, some unique and varied approaches have achieved state-of-the-art results with relatively fast run times~\cite{balakrishnan2019voxelmorph,li2017non,miao2016cnn,yang2017quicksilver}. Li et al.~\cite{li2017non} propose a fully convolutional “self-supervised” approach to learn the appropriate spatial transformations at multiple resolutions. Balakrishnan et al.~\cite{balakrishnan2019voxelmorph} propose a method for unsupervised image registration which attempts to directly compute the deformation field.

\underline{Neuroimaging prediction}. With many architectures being borrowed from the computer vision community, deep learning based prediction in neuroimaging has quickly gained popularity. Traditionally, machine learning based prediction on neuroimaging data has relied on careful feature selection/engineering; often taking the form of regional summary measures which may not account for all the informative variation for a particular task. Whereas, in deep learning, it is common to work with raw imaging data, where the appropriate feature representations can be learned through optimization. This can be particularly useful for high-level prediction tasks in which we do not know what imaging features will be informative. Further, by working on the raw image, reliance on complex and time consuming preprocessing can be reduced. In recent years, a large amount of work has been published on deep learning based prediction tasks such as brain age prediction~\cite{bashyam2020mri,cole2017predicting,jonsson2019brain}, Alzheimer’s disease classification and trajectory modeling~\cite{liu2014early,lu2018multimodal,islam2018brain,liu2018landmark}, and schizophrenia classification~\cite{yan2019discriminating,zeng2018multi}. Some work has considered the use of deep Siamese networks for longitudinal image analysis. Siamese networks have gained popularity for their success in facial recognition. They work by jointly optimizing a set of weights on two images  with respect to some distance metric between them. This setup makes them effective at identifying longitudinal  changes on some chosen dimension. Bhagwat et al.~\cite{bhagwat2018modeling} consider the use of longitudinal Siamese network for the prediction of future Alzheimer’s disease onset, using two early time points. They show substantially improved performance in identifying future Alzheimer’s cases, with the use of two time points versus only a baseline scan. 

\underline{The use of GANs in neuroimaging}.
GANs have enabled complex image synthesis tasks in neuroimaging, many of which have no comparable analogs in traditional machine learning. GANs and their variants have been used in neuroimaging for cross-modality synthesis~\cite{wolterink2017deep}, motion artifact reduction~\cite{tamada2018motion}, resolution upscaling~\cite{chen2018brain,pham2017brain,song2020super}, estimating full-dosage PET images from low-dosage PET~\cite{chen2019ultra,xu2017200x,kaplan2019full}, image harmonization~\cite{dewey2019deepharmony,nath2019inter}, heterogeneity analysis~\cite{yang2020smile}, and more. To help facilitate such work, the popular MedGAN~\cite{armanious2020medgan} proposes a series of modifications and new loss functions to traditional GANs, aimed at preserving anatomically relevant information and fine details. They use auxiliary classifiers on the translated image to ensure the resulting image feature representation is similar to the expected image representation for a given task. Additionally, they use style-transfer loss in combination with adversarial loss to ensure fine structures and textural details are matched in the translation. Some promising new work attempts to reduce the amount of radioactive tracer needed in PET imaging, potentially reducing associated costs and health risks. This problem can be framed as an image synthesis task, whereby the final image can be synthesized from the low dose image. In~\cite{van2015cross}, the pixel location information is integrated into a deep network for image synthesis. Kaplan and Zhu~\cite{kaplan2019full} propose a deep generative based denoising method, that uses paired scans of subjects imaged with both low and full dose PET. They show that despite a ten-fold reduction in tracer material, they are able to preserve important edge, structural, and textural details. Consistent quantification in neuroimaging has been hampered for decades by the high variability in MR image intensities and resolutions between scans. Dewey et al.~\cite{dewey2019deepharmony} use a U-Net style architecture and paired subjects who have been scanned with two different protocols, to learn a mapping between the two sites. Resolution differences are addressed by applying a super-resolution method to the images acquired at lower resolutions~\cite{zhao2018deep}. They are able to use the network to reduce site based variation, which improves consistency in segmentation between the two sites.

While deep learning in neuroimaging has certainly opened up many interesting avenues of investigation, certain areas still lack a rigorous understanding. Important lines of research such as learning from limited data, optimal hyperparameter selection, domain adaptation, semi-supervised designs, and improving robustness require further investigation.

\subsection{Deep learning in cardiovascular imaging} 

The quantification and understanding of cardiac anatomy and function has been transformed by the recent progress in the field of data-driven deep learning. 
There has been significant recent work in a variety of sub-areas of cardiovascular imaging 
including image reconstruction~\cite{Bustin2020}, end-to-end learning of cardiac pathology from images~\cite{Zheng2019} and incorporation of non-imaging information (e.g. genetics~\cite{deMarvao2020} and clinical information) for analysis. Here we briefly focus on three key aspects of deep learning in this field: cardiac chamber segmentation, cardiac motion/deformation analysis and analysis of cardiac vessels. 
Motion tracking and segmentation both play crucial roles in the detection and quantification of myocardial chamber dysfunction and can help in the diagnosis of cardiovascular disease (CVD). Traditionally, these tasks are treated uniquely and solved as separate steps. Often times, motion tracking algorithms use segmentation results as an anatomical guide to sample points and regions of interest used to generate displacement fields 
~\cite{PARAJULI2019116}. In part due to this, there have also been efforts to combine motion tracking and segmentation.


\underline{Cardiac image segmentation} is an important first step for many clinical applications. The aim is typically to segment the main chambers, e.g. the left ventricle (LV), right ventricle (RV), left atrium (LA), and right atrium (RA). This enables the quantification of parameters that describe cardiac morphology, e.g. LV volume or mass, or cardiac function, e.g. wall thickening and ejection fraction. 
There has been significant deep learning work on cardiac chamber segmentation, mostly characterized by the type of images (modalities) employed and whether the work is in 2D or 3D. One of the first efforts to apply a fully convolutional network (FCN)~\cite{Shelhamer2017} to segment the left ventricle (LV),  myocardium and right ventricle from 2D short-axis cardiac magnetic resonance (MR) images was done by Tran~\cite{Tran2016}, significantly outperforming traditional methods in accuracy and speed.
Since this time, a variety of other FCN-based strategies have been developed~\cite{Bai_2018_JCMR}, especially focusing on the popular U-Net approach, often including both 2D and 3D constraints (e.g.~\cite{Isensee2017}). The incorporation of spatial and temporal context has also been an important research direction, including efforts to simultaneously segment the heart in both the end-diastolic and end-systolic states~\cite{Wolterink2017}. Shape-based constraints had previously been found useful for LV chamber segmentation using other types of machine learning (e.g.~\cite{Huang2014}) and were nicely included in an anatomically-constrained deep learning strategy by~\cite{Oktay2018}. This stacked convolutional autoencoder approach was successfully applied to LV segmentation from 3D echocardiography data as well.  Other important work has been aimed at atrial segmentation from MRI~\cite{Xiong2019}, whole heart segmentation from CT~\cite{Ye2019} and LV segmentation from 3D ultrasound image sequences~\cite{Dong2018}, the latter using a combination of atlas registration and adversarial learning. Progress in deep learning for cardiac segmentation is enabled by a number of ongoing challenges in the field ~\cite{stacom2019,stacom2020}.


\begin{figure}[t]
\centering
    \includegraphics[width=1.0\columnwidth]{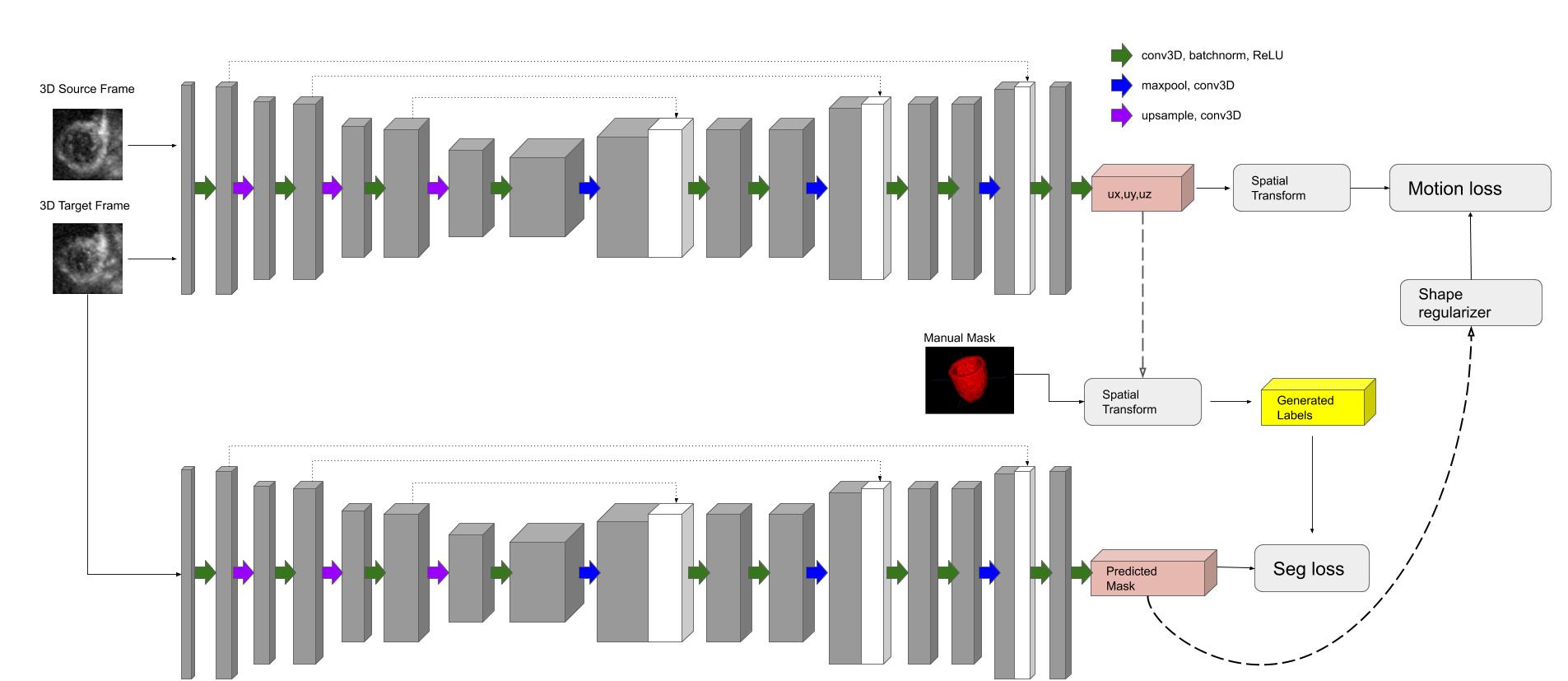} \\ 
        \includegraphics[width=1.0\columnwidth]{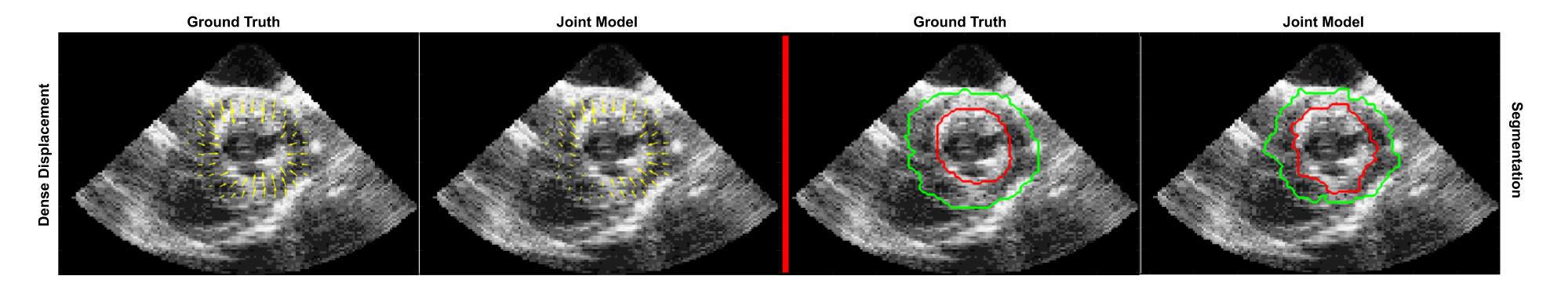}
        \caption{Top: 4D two-channel CNN architecture for  joint LV motion tracking/segmentation network; Bottom:  2D echocardiography slices through 3D canine results (left = displacements with ground truth interpolated from sonomicrometer crystals; right = LV endocardial (red) and epicardial (green) segmented boundaries with ground truth from human expert tracing.)  Reproduced from \cite{Ta2020}. } 
        \label{fig:cardiac-joint}
\end{figure}


\underline{Cardiac motion tracking} is key for deformation/strain analysis and is important for analyzing the mechanical performance of heart chambers. A variety of image registration, feature-based tracking and regularization methods using both biomechanical models and data-driven learning have been developed. One special type of dataset useful for tracking are MRI tagging acquisitions, and deep learning has recently played a role in tracking these tags and quantifying the displacement information for motion tracking and analysis~\cite{Ferdian2020} using a combination of recurrent neural networks (RNNs) and convolutional neural networks (CNNs) to estimate myocardial strain from short axis MRI tag image sequences. Estimating motion displacements and strain is also possible to do
from both standard MR image sequences and 4D echocardiography, most often by integrating ideas of image segmentation and mapping between frames using some type of image registration.
%
Recent efforts for cardiac motion tracking from magnetic resonance (MR) imaging have adopted approaches from the computer vision field, suggesting that the tasks of motion tracking and segmentation are closely related and information used to complete one task may complement and improve the overall performance of the other. In particular, an interesting deep learning approach proposed for joint learning of video object segmentation and optical flow (motion displacements) is termed
SegFlow~\cite{Cheng_ICCV_2017}, 
an end-to-end unified network that simultaneously trains both tasks and exploits the commonality of these two tasks through bi-directional feature sharing. 
Among the first to integrate this idea into cardiac analysis was Qin et al.~\cite{qin2018joint}, who successfully implemented the idea of combining motion and segmentation on 2D cardiac MR sequences by developing a dual Siamese style recurrent spatial transformer network and fully convolutional segmentation network to simultaneously estimate motion and generate segmentation masks. 
This work was mainly aimed at 2D MR images, which have higher SNR than echocardiographic images and, therefore, more clearly delineated LV walls. It remains challenging to directly apply this approach to echocardiography. 
Very recent efforts by Ta et al.~\cite{Ta2020} (see Fig.~\ref{fig:cardiac-joint}) propose a 4D (3D+t) semi-supervised joint network to simultaneously track LV motion while segmenting the LV wall. The network is trained in an iterative manner where results from one branch influence and regularize the other. Displacement fields are further regularized by a biomechanically-inspired incompressibility constraint that enforces realistic cardiac motion behavior. The proposed model is different from other models in that it expands the network to 4D in order to capture out of plane motion. Finally, clinical interpretability of deep learning-derived motion information will be an important topic in the years ahead (e.g.~\cite{Bello2019}).



\underline{Cardiac {vessel segmentation}} is another important task for cardiac image analysis and includes the segmentation of the vessels including the great vessels (e.g. aorta, pulmonary arteries and veins) as well as the coronary arteries. The segmentation of large vessels such as the aorta is important for accurate mechanical and hemodynamic characterization, e.g. for assessment of aortic compliance. Several deep learning approaches have been proposed for this segmentation task, including the use of recurrent neural networks in order to track the aorta in cardiac MR image sequences in the presence of noise and artefacts~\cite{Bai_2018_MICCAI}. A similarly important task is the segmentation of the coronary arteries as a precursor to quantitative analysis for the assessment of stenosis or the simulation of blood flow simulation for the calculation of fractional flow reserve from CT angiography (CTA). The approaches for coronary artery segmentation can be divided into those approaches that extract the vessel centerline and those that segment the vessel lumen. 

One end-to-end trainable approach for the extraction of the coronary centerline has been proposed in~\cite{Guo_2019_IPMI}. In this approach the centerline is extracted using a multi-task fully convolutional network which simultaneously computes centerline distance maps and detects branch endpoints. The method generates single-pixel-wide centerlines with no spurious branches. An interesting aspect of this technique is that it can handle an arbitrary vessel tree with no prior assumption regarding depth of the vessel tree or its bifurcation pattern. In contrast to this, Wolterink et al.~\cite{Wolterink_2019_MedIA} propose a CNN that is trained to predict the most likely direction and radius of the coronary artery within a local 3D image patch. Starting from a seed point, the coronary artery is tracked by following the vessel centerline using the predictions of the CNN. 

Alternative approaches to centerline extraction are based on techniques that instead aim to segment the vessel lumen, e.g. using CNN segmentation methods that perform segmentation by predicting vessel probability maps. One elegant approach has been proposed by Moeskops et al.~\cite{Moeskops_2016_MICCAI}: Here a single CNN is trained to perform three different segmentation tasks including coronary artery segmentation in cardiac CTA. Instead of performing voxelwise segmentation, Lee et al.~\cite{Lee_2019_TMI} introduce a tubular shape prior for the vessel segments. This is implemented via a template transformer network, through which a shape template can be deformed via network-based registration to produce an accurate segmentation of the input image, as well as to guarantee topological constraints. 

More recently, geometric deep learning approaches have also been applied for coronary artery segmentation. For example, Wolterink et al.~\cite{Wolterink2019-graph} used graph convolutional networks for coronary artery segmentation. Here vertices on surface of the coronary artery are used as graph nodes and their locations are optimized in an end-to-end fashion.

\subsection{Deep learning in abdominal imaging} 
Recently there has been an accelerating progress in automated detection, classification and segmentation of abdominal anatomy and disease using medical imaging. Large public data sets such as the MICCAI Data Decathlon and Deep Lesion data sets have facilitated progress~\cite{simpson2019large,yan2018deeplesion}.

\begin{figure}[t]
    \centering
    \includegraphics[width=\columnwidth]{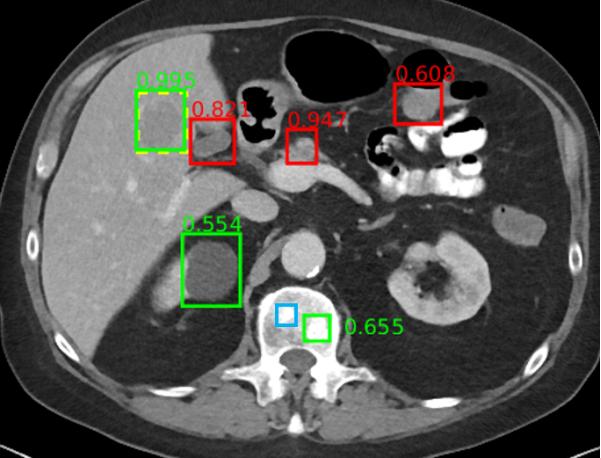}
    \caption{Example universal lesion detector for abdominal CT. In this axial image through the upper abdomen, a liver lesion was correctly detected with high confidence (0.995). A renal cyst (0.554) and a bone metastasis (0.655) were also detected correctly. False positives include normal pancreas (0.947), gallbladder (0.821), and bowel (0.608). A subtle bone metastasis (blue box) was missed. Reproduced from \cite{yan2018deeplesion}.}
    \label{fig:lesion}
\end{figure}

\underline{Organs and lesions.} Multi-organ approaches have been popular methods for anatomy localization and segmentation~\cite{cerrolaza2019computational}. For individual organs, the liver, prostate and spine are arguably the most accurately segmented structures and the most actively investigated with deep learning. Other organs of interest to deep learning researchers include pancreas, lymph nodes and bowel. 

A number of studies have used U-Net to segment the liver and liver lesions and assess for hepatic steatosis~\cite{seo2019modified,graffy2019automated,tang2020e2net}. Dice coefficients for liver segmentation typically exceed 95\%. In the prostate, gland segmentation and lesion detection have been the subject of an SPIE/AAPM challenge (competition) and numerous publications~\cite{armato2018prostatex,cheng2019fully}. 
Several groups have used data sets such as TCIA CT pancreas to improve pancreas segmentation with Dice coefficients reaching the mid 80 percentile~\cite{roth2018spatial,roth2015data,roth2018application,chu2019application}. Automated detection of pancreatic cancer using deep learning has also been reported~\cite{chu2019application}. Deep learning has been used for determining pancreatic tumor growth rates in patients with neuroendocrine tumors of the pancreas~\cite{zhang2019spatio}. The spleen has been segmented with a Dice score of 0.962~\cite{humpire2020fully}. Recently marginal loss and exclusion loss~\cite{shi2020marginal} have been proposed to train a single multi-organ segmentation network from a union of partially labelled datasets.

Enlarged lymph nodes can indicate the presence of inflammation, infection, or metastatic cancer. Studies have assessed abdominal lymph nodes on CT in general and for specific diseases such as prostate cancer ~\cite{roth2016improving,shin2016deep,tang2020one}. The TCIA CT lymph node dataset has enabled progress in this area~\cite{roth2015data2}.

In the bowel, CT colonography computer-aided polyp detection was a hot topic in abdominal CT image analysis over a decade ago. Recent progress with deep learning has been limited but studies have reported improved electronic bowel cleansing, and higher sensitivities and lower false-positive rates for precancerous colonic polyp detection~\cite{roth2016improving,tachibana2018deep}. Deep learning using persistent homology has recently shown success for small bowel segmentation on CT ~\cite{Shin2020deep}. Colonic inflammation can be detected on CT with deep learning~\cite{liu2017detection}. Appendicitis can be detected on CT scans by pre-training with natural world videos~\cite{rajpurkar2020appendixnet}. The Inception V3 convolutional neural network could detect small bowel obstruction on abdominal radiographs~\cite{cheng2018detection}.

Kidney function can be predicted using deep learning of ultrasound images~\cite{kuo2019automation}. Potentially diffuse disorders such as ovarian cancer and abnormal blood collections were detectable using deep learning~\cite{wang2019deep,dreizin2020performance}. Organs at risk for radiation therapy of the male pelvis, such as bladder and rectum, have been segmented on CT using U-Net~\cite{kazemifar2018segmentation}.

Universal lesion detectors~\cite{yan2018deeplesion,li2020bounding} have been developed for body CT including abdominal CT (Figure~\ref{fig:lesion}). The universal lesion detector identifies, classifies and measures lymph nodes and a variety of tumors throughout the abdomen. This detector was trained using the publicly available Deep Lesion data set. 

\underline{Opportunistic screening} to quantify and detect under-reported chronic diseases has been an area of recent interest. Example deep learning methods for opportunistic screening in the abdomen include automated bone mineral densitometry, visceral fat assessment, muscle volume and quality assessment, and aortic atherosclerotic plaque quantification~\cite{pickhardt2020automated}. Studies indicate that these measurements can be done accurately and generalize well to new patient populations. These opportunistic screening assessments also enable prediction of survival and cardiovascular morbidity such as heart attack and stroke~\cite{pickhardt2020automated}.

Deep learning for abdominal imaging is likely to continue to advance rapidly. For translation to the clinic, some of the most important advances sought will be in demonstrating generalizability across different patient populations and variations in image acquisition.

\subsection{Deep learning in microscopy imaging} 

With the advent of whole slide scanning and the development of large digital datasets of tissue slide images, there has been a significant increase in application of deep learning approaches to digital pathology data~\cite{bera2019artificial}. While the initial application of these approaches in the area of digital pathology primarily focused on its utility for detection and segmentation of individual primitives like lymphocytes and cancer nuclei, they have now progressed to addressing higher level diagnostic and prognostic tasks and also the application of DL approaches to predict the underlying molecular underpinning and mutational status of the disease. Briefly below we describe the evolving applications of DL approaches to digital pathology.

\underline{Nuclei detection and segmentation}. One of the early applications of DL to whole slide pathology images was in the detection and segmentation of individual nuclei. Xu et al.~\cite{xu2015stacked} present an approach using stacked spare auto-encoder approach to identify the location of individual cancer nuclei on breast cancer pathology images. Subsequently work from Janowczyk et al.~\cite{janowczyk2016deep} demonstrate the utility of DL approaches for identifying and segmenting a number of different histologic primitives including lymphocytes, tubules, mitotic figures, cancer extent and also for classifying different disease categories pertaining to leukemia. The comprehensive tutorial also went into great detail with regard to best practices for annotation, network training and testing protocols. Subsequently Cruz-Roa et al. demonstrate that convolutional neural networks could be applied for accurately identifying cancer presence and extent on whole slide breast cancer pathology images~\cite{cruz2017accurate}. The approach is shown to have a 100\% accuracy at identifying the presence or absence of cancer on a slide or patient level. Subsequently Cruz-Roa et al also demonstrate the use of a high-throughput adaptive sampling approach for improving the efficiency of the CNN presented previously in~\cite{cruz2018high}. In~\cite{bejnordi2017diagnostic} Veta and colleagues discuss the diagnostic assessment of DL algorithms for detection of lymph node metastases in women with breast cancer, as part of the CAMELYON16 challenge. The work find that at least five DL algorithms perform comparably to a pathologist interpreting the slides in the absence of time constraints and that some DL algorithms achieve better diagnostic performance than a panel of eleven pathologists participating in a simulation exercise designed to mimic a routine pathology workflow. In a related study of lung cancer pathology images, Coudray et al.~\cite{coudray2018classification} train a deep CNN (inception V3) on WSIs from The Cancer Genome Atlas (TCGA) to accurately and automatically classify them into lung adenocarcinoma, squamous cell carcinoma or normal lung tissue, yielding an area under the curve of 0.97.

\begin{figure}[t]
\centering
    \includegraphics[width=\columnwidth]{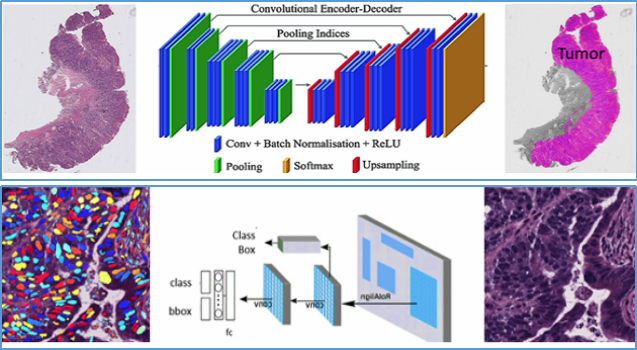}
    \caption{Application of deep learning for identifying cancerous regions from whole slide images as well as for identifying and segmenting different types of nuclei within the whole slide pathology images.} 
        \label{fig:microscopy_appl}
\end{figure}

One of the challenges with the CNN based approaches described in~\cite{cruz2017accurate,cruz2018high,bejnordi2017diagnostic,coudray2018classification} is the need for detailed annotations of the target of interest. This is a labor intensive task given that annotations of disease extent typically need to be provided by pathologists who have a minimal amount of time to begin with. In a comprehensive paper by Campanella et al.~\cite{campanella2019clinical}, the team employs a weakly supervised approach for training a deep learning algorithm for identifying the presence or absence of cancer on a slide level. They are able to demonstrate on a large scale study of over 44K WSIs from over 15K patients, that for prostate cancer, basal cell carcinoma and breast cancer metastases to axillary lymph nodes, the corresponding areas under the curve are all above 0.98. The authors suggest that the approach could be used by pathologists to exclude 65-75\% of slides while retaining 100\% cancer detection sensitivity.

\underline{Disease grading}. 
Pathologists can reliably identify disease type and extent on H\&E slides and have made observations for decades that there are features of disease that correlate with its behavior. However they are unable to reproducibly identify or quantify these histologic hallmarks of disease behavior with enough rigor, to use these features routinely to dictate disease outcome and treatment response. One of the areas to which DL has been applied is mimicking pathologist’s identification of disease hallmarks, especially in the context of cancer. For instance in prostate cancer, pathologists typically aim to place the cancer into one of five different categories, referred to as the Gleason grade groups~\cite{babaian1985reliability}. However this grading system, as with many other cancers and diseases, is subject to inter-reader variability and disagreement. Consequently, a number of recent DL approaches for prostate cancer grading have been presented. Bulten et al.~\cite{bulten2020automated} and Strom et al.~\cite{strom2020artificial} both recently publish large cohort studies involving 1,243 and 976 patients, respectively, and show that DL approaches could be used for achieving a performance of Gleason grading that is comparable to pathologists~\cite{madabhushi2020deep}. 

\underline{Mutation identification and pathway association}. Disease morphology reflects the sum of all temporal genetic and epigenetic changes and alterations within the disease. Recognizing this, some groups have begun to explore the role of DL approaches for identifying disease specific mutations and associations with biological pathways. Oncotype DX is a 21-gene expression assay that is prognostic and predictive of benefit of adjuvant chemotherapy for early stage estrogen receptor positive breast cancers. In two related studies~\cite{romo2016automated,romo2017deep}, Romo-Bucheli show that DL could be used to identify tubule density and mitotic index from pathology images and demonstrate a strong association between these measurements with the Oncotype DX risk categories (low, intermediate, and high) for breast cancers. Interestingly, tubule density and mitotic index are an important component of breast cancer grading. Microsatellite instability (MSI) is a condition that results from impaired DNA mismatch repair. To assess whether a tumor is MSI, genetic or immunohistochemical tests are required. A study by Kather et al.~\cite{kather2019deep} shows that DL could predict MSI from histology images in gastrointestinal cancers with an AUC=0.84. Coudray et al.~\cite{coudray2018classification} show that a DL network can be trained to recognize many of the commonly mutated genes in non-small cell lung adenocarcinoma. They show that six of these mutated genes---TK11, EGFR, FAT1, SETBP1, KRAS, and TP53---can be predicted from pathology images, with AUCs from 0.733 to 0.856.

\underline{Survival and disease outcome prediction}.
More recently, there has been an interest in applying DL algorithms to pathology images to directly predict survival and disease outcome. In a recent paper, Skrede et al.~\cite{skrede2020deep} perform a large study of DL involving over 12M pathology image tiles from over 2,000 patients to predict cancer specific survival in early stage colorectal cancer patients. DL yields a hazard ratio for poor versus good prognosis of 3.84 (95\% CI 2.72-5.43; p$<$0.0001) in a validation cohort of 1,122 patients and a hazard ratio of 3.04 (2.07-4.47; p$<$0.0001) after adjusting for established prognostic markers including T and N stages. Courtiol et al.~\cite{courtiol2019deep} present an approach employing DL for predicting patient outcome in the case of mesothelioma. Saillard et al.~\cite{saillard2020predicting} use DL to predict survival after hepatocellular carcinoma resection. 

While the studies described above clearly reflect the growing influence and impact of DL on a variety of image analysis and classification problems in digital pathology, there are still concerns with regards to its interpretability, the need for large training sets, the need for annotated data, and generalizability. Attempts have been made to use approaches like visual attention mapping~\cite{tomita2019attention} to provide some degree of transparency with respect to where in the image the DL network appears to be focusing its attention. Another approach to imbue interpretability is via hybrid approaches, wherein DL is used to identify specific primitives of interest (e.g. lymphocytes) in the pathology images (in other words using it as a detection and segmentation tool) and then deriving hand-crafted features from these primitives (e.g. spatial patterns of arrangement of lymphocytes) to perform prognosis and classification tasks~\cite{heindl2018relevance,saltz2018spatial,corredor2019spatial}. However, as Bera et al.~\cite{bera2019artificial} note in a recent review article, while DL approaches might be feasible for diagnostic indications, clinical tasks relating to outcome prediction and treatment response might still involve approaches that provide greater interpretability.  While it seems highly likely that research in DL and its application to digital pathology are likely to continue to grow, it remains to be seen how these approaches fare in a prospective and clinical trial setting, which in turn might ultimately determine their translation to the clinic.

\section{Discussion} 

{\underline{Technical challenges ahead.}}
In this overview paper, many technological challenges across several medical domains and tasks have been reviewed.  In general, most challenges are met by continuous improvement of solutions to the well known {\em{data challenge}}. The community as a whole is continuously developing and improving transfer learning based solutions and data augmentation schemes. As systems are starting to be implemented across datasets, hospitals, and countries, a new spectrum of challenges is arising including {\em{system robustness and generalization}} across acquisition protocols, machines, and hospitals. Here, data pre-processing, continuous model learning, and fine-tuning across systems are a few of the new developments ahead. Detailed reviews of the topics presented herein, as well as additional topics, such as robustness to adversarial
attacks~\cite{yao2020hierarchical}, can be found in several recent DL review articles such as \cite{R_S_2020}.

{\underline{How do we get new tools into the clinic?}}
The question whether DL tools are used in the clinic is often raised~\cite{recht2020integrating}. This question is particularly relevant because results in many tasks and challenges show radiologist-level performance. In several recent works conducted to estimate the utility of AI-based technology as an aid to the radiologist, it is consistently shown that human experts with AI perform better than those without AI~\cite{Brown2019IntegrationOC}. The excitement in the field has led to the emergence of many AI medical imaging startup companies. Although there are some earlier studies ~\cite{de2010computer,jorritsma2015improving,drew2012and} that present evidence that early CAD tools were not necessarily helpful in the tested scenarios and, to date, the translation of technologies from research to actual clinical use does not happen at a massive scale, there are a number of technologies that obtained the FDA approval\footnote{https://www.acrdsi.org/DSI-Services/FDA-Cleared-AI-Algorithms}. 
In fact, a dedicated reimbursement code was recently awarded for an AI technology\footnote{Federal Register / Vol. 85, No. 182 / Friday, September 18, 2020 / Rules and Regulations}. There are a variety of reasons for this delayed clinical translation including: users being cautious regarding the technology, specifically the prospect of being replaced by AI;  the need to prove that the technology can address real user needs and bring quantifiable benefits; regulatory pathways that are long and costly; patient safety considerations; and economic factors such as who will pay for AI tools. 

The forecast going forward is that this is an emerging field, with enormous promise going forward. How would we get there? An interesting possibility is that the worldwide experience with the COVID-19 pandemic will actually serve to bridge the gap between the {\em {need}} and {\em{AI}}---with users eager to receive support, and even the regulatory steps more adaptive to facilitate a transition of general computational tools, with COVID-AI related tools in particular. Within the last several months, we witnessed several interesting advances: the ability of AI to rapidly adapt from existing pretrained models to new disease manifestations of COVID-19, using the many tools described in this article and in~\cite{gozes2020rapid,murphy2020covid}. Strong and robust DL based solutions for COVID-19 detection, localization, quantification, and characterization start to support the initial diagnosis and more so the followup of hospitalized patients. AI-based tools are developed to support the assessment of disease severity and, recently, they start to provide tools for assessing treatment and predicting treatment success~\cite{Dinggang,Greenspan_MedIA_2020}. Finally, numerous studies ~\cite{RATHORE2017530,mateos2018structural,nielsen2019machine,fathi2020imaging}
in fields like clinical neuroscience have shown that AI-based image evaluation can identify complex imaging patterns that are not perceptible with visual radiologic evaluation. For example anatomical, functional, and connectomic signatures of many neusopsychiatic and neurologic diseases not only promise to redefine many of these diseases on a neurobiological basis and in a more precise manner, but also to offer early detection and personalized risk estimates. Such capabilities can potentially significantly expand current clinical protocols and contribute to precision medicine.

{\underline{Future promise}.}
As we envision future possibilities, one immediate step forward is to combine the image with additional clinical context, from patient record to additional clinical descriptors (such as blood tests, genomics, medications, vital signs, and non-imaging data such as ECG). This step will provide a transition from {\em{image space }} to {\em{ patient-level}} information. 
Collecting cohorts will enable population-level statistical analysis to learn about disease manifestations, treatment responses, adverse reactions from and interactions between medications, and more.  This step requires building complex infrastructure, along with the generation of new privacy and security regulations---between hospitals and academic research institutes, across hospitals, and in multi-national consortia.
As more and more data become available, DL and AI will enable unsupervised explorations within the data, thus providing for new discoveries of drugs and treatments towards  advancement and augmentation of healthcare as we know it. 


%

\section*{Acknowledgement}
Ronald M. Summers was supported in part by the National Institutes of Health Clinical Center. Anant Madabhushi thanks research support by the National Institutes of Health under award numbers 1U24CA199374-01, R01CA202752-01A1, R01CA208236-01A1, R01CA216579-01A1, R01CA220581-01A1, 1U01CA239055-01, 1U54CA254566-01, 1U01CA248226-01, 1R43EB028736-01 and the VA Merit Review Award IBX004121A from the United States Department of Veterans Affairs Biomedical Laboratory Research and Development Service. We extend our special acknowledgement to Vishnu Bashyam for his help. The mentioning of commercial products does not imply endorsement.

\bibliographystyle{IEEEtran}
\bibliography{bib_dlmi,bib_neuro,bib_cardiac,bib_abdomen,bib_microscopy,bib_chest}
\end{document}